\documentclass{article}

\PassOptionsToPackage{numbers, compress}{natbib}

\usepackage[final]{neurips_2020}

\usepackage[utf8]{inputenc} 
\usepackage[T1]{fontenc}    
\usepackage{hyperref}       
\usepackage{url}            
\usepackage{booktabs}       
\usepackage{amsfonts}       
\usepackage{nicefrac}       
\usepackage{microtype}      
\usepackage{graphicx}
\usepackage{multirow}
\usepackage{subfigure}
\usepackage{multicol}
\usepackage{float}

\title{Hi-UCD: A Large-scale Dataset for Urban Semantic Change Detection in Remote Sensing Imagery}

\author{%
  Shiqi Tian \\
  \And
   Ailong Ma 
  \And
  Zhuo Zheng
  \And
  Yanfei Zhong   \thanks{Corresponding author.}
  \AND
  Wuhan University \\
  Wuhan, China \\
  \texttt{\{shiqitian,maailong007,zhengzhuo,zhongyanfei\}@whu.edu.cn} 
}

\begin{document}

\maketitle

\begin{abstract}
  With the acceleration of the urban expansion, urban change detection (UCD), as a significant and effective approach, can provide the change information with respect to geospatial objects for dynamical urban analysis. However, existing datasets suffer from three bottlenecks: (1) lack of high spatial resolution images; (2) lack of semantic annotation; (3) lack of long-range multi-temporal images. In this paper, we propose a large scale benchmark dataset, termed Hi-UCD. This dataset uses aerial images with a spatial resolution of 0.1 m provided by the Estonia Land Board, including three-time phases, and semantically annotated with nine classes of land cover to obtain the direction of ground objects change. It can be used for detecting and analyzing refined urban changes. We benchmark our dataset using some classic methods in binary and multi-class change detection. Experimental results show that Hi-UCD is challenging yet useful. We hope the Hi-UCD can become a strong benchmark accelerating future research.   
\end{abstract}

\section{Introduction}
Change detection obtains ground feature change information by comparing images from different periods.
Remote sensing images have become common data for detecting changes in the surface due to their high spatial coverage and high time resolution \cite{Hussain.2013}. At the same time, the increased spatial resolution of remote sensing images can provide more details of ground objects.
In-depth study of urban change is essential to promote sustainable urbanization \cite{Huang.2017}. Therefore, UCD has become a research hotspot. 
Urban areas often have a wide variety of objects and strong regional heterogeneity. Ground objects, even in the same class, may have very different geometric shapes, and local features. In order to better analyze urbanization, different requirements are also put forward for the usage data.
\textbf{(1) Higher spatial resolution.}
 Higher spatial resolution images can provide more information to distinguish features between different images to obtain a clear boundary of change.
\textbf{(2) Richer prior information on land cover.} Knowing the prior information about land cover can detect the direction of change and analyze land cover changes.
\textbf{(3) Longer time series images.} 
The changes in ground objects are time-dependent, and a more continuous sequence of images can realize time series analysis to monitor urban changes.

We collected public UCD datasets (Table~\ref{tab:dataset}) and found that they have some limitations: 
\textbf{(1) Lack of high spatial resolution images.} The image spatial resolutions of OSCD \cite{Daudt.2018}, ZY3 \cite{Zhang.2020}, and SZTAKI AirChange \cite{Benedek.2009} are 10 m, 5.8 m, and 1.5 m respectively. Although the resolution is gradually increasing, it still cannot meet the requirements of UCD, especially for buildings.
\textbf{(2) Lack of semantic annotation.} ABCD \cite{Fujita.2017}, LEVIR-CD \cite{Chen.2020}, WHU Building \cite{Ji.2019b} only label building-related changes, and Season-varying \cite{Lebedev.2018} directly labeled land cover related changes, all of them are lack of semantic changes. It is difficult to perform multi-class change detection to obtain fine change directions. Although HRSCD \cite{CayeDaudt.2019} provided the direction of changes, its labeling accuracy is only
80\% to 85\%. In addition, the ground objects of the urban area are classified into five categories,
which is relatively rough and difficult to reflect the changes of typical objects in urban areas.
\textbf{(3) Lack of long-range multi-temporal images.} The above-mentioned public datasets only contain bi-temporal images of the same area. Therefore, it is difficult to obtain satisfactory refined detection results for UCD.

\begin{table*}[!t]
  \centering
  \caption{ Open datasets in remote sensing change detection } \label{tab:dataset}
    \resizebox{\linewidth}{!}{
    \begin{tabular}{cccccccc}
      \toprule
      {Dataset}& Resolution (m) & \#Images & Image size (Pixel)  & Years & Change &Interesting object& Classes \\
      \midrule

      OSCD \cite{Daudt.2018}  & 10 & 24 & $600\times600$  & 2 & Binary change & Land cover&- \\
    
      ZY3 \cite{Zhang.2020}  & 5.8 & 1 & $1154\times740$  & 2  & Binary change & Land cover&- \\

       SZTAKI AirChange \cite{Benedek.2009}  & 1.5 & 13 & $952\times640$  & 2 & Binary change & Land cover&-\\

       AICD \cite{Bourdis.2011}  & 0.5 & 1000 & $800\times600$  & 2 & Binary change & Land cover&-\\
      
       {ABCD\cite{Fujita.2017}}  & {0.4} & 8506/8444 & $160\times160$/$120\times120$  &{2} &Binary change & {Building} &-\\

        LEVIR-CD \cite{Chen.2020}  & 0.5 & 637 & $1024\times1024$  & 2 & Binary change & Building&-\\

       WHU Building \cite{Ji.2019b}& 0.2 & 1 & $32207\times15354$  & 2  & Binary change & Building&-\\

      Season-varying \cite{Lebedev.2018}&0.03-1&16000& $256\times256$&2&Binary change&Land cover&-\\
      HRSCD \cite{CayeDaudt.2019}  & 0.5 & 291 & $10000\times10000$  & 2 & Semantic change & Land cover&5 \\
      Hi-UCD (ours)  & 0.1 & 1293 & $1024\times1024$  & 3 & Semantic change & Land cover & 9 \\
      \bottomrule
    \end{tabular}}

\end{table*}

To solve these problems, we introduce a large-scale semantic annotated ultra-high resolution UCD dataset named Hi-UCD.
Our dataset uses aerial images with a spatial resolution of 0.1m to clearly show the spatial details of ground objects and capture small changes in them.
Hi-UCD obtains fine semantic changes of objects by labeling the land cover classes of images in different periods. We have selected 9 land cover classes including natural and artificial objects to achieve full coverage of urban ground objects.
In addition, Hi-UCD contains images of three time phases in the same area, which is conducive to studying the temporal correlation of changes in ground features.
Overall, Hi-UCD is a large-scale, multi-temporal, ultra-high resolution urban semantic change detection data set, which can realize comprehensive detection and analysis of urban changes.
To verify the validity of Hi-UCD, we selecte the classic method in the binary and multi-class change detection task to conduct the experiments, finally provide a benchmark.

\section{Hi-UCD Dataset}

Hi-UCD focuses on urban changes and uses ultra-high resolution images to construct multi-temporal semantic changes to achieve refined change detection.
The study area of Hi-UCD is a part of Tallinn, the capital of Estonia, with an area of 30 $km^2$. The Estonian Land Board provides aerial images\footnote{ Orthophoto, Land Board 2020.} taken by Leica ADS100-SH100 in 2017, 2018, and 2019, with topographic database\footnote{Estonian Topographic Database, Land Board 2020.} for the area. Hi-UCD obtained semantic changes by annotating the land cover classes in different periods. We have considered topographic documents and changes in ground objects to
select 9 land cover classes to achieve complete coverage of ground objects in Estonia.
 Finally, we cut each year's images into patches with a size of $1024\times1024$, and filter out patches with change pixels more than 200 to form the Hi-UCD dataset. There are 359 image pairs in 2017-2018, 386 pairs in 2018-2019, and 548 pairs in 2017-2019, including images, semantic maps and change maps at different times.
\begin{figure}[!ht]
  \centering
  \includegraphics[scale=0.4,trim=0 140 150 0,clip]{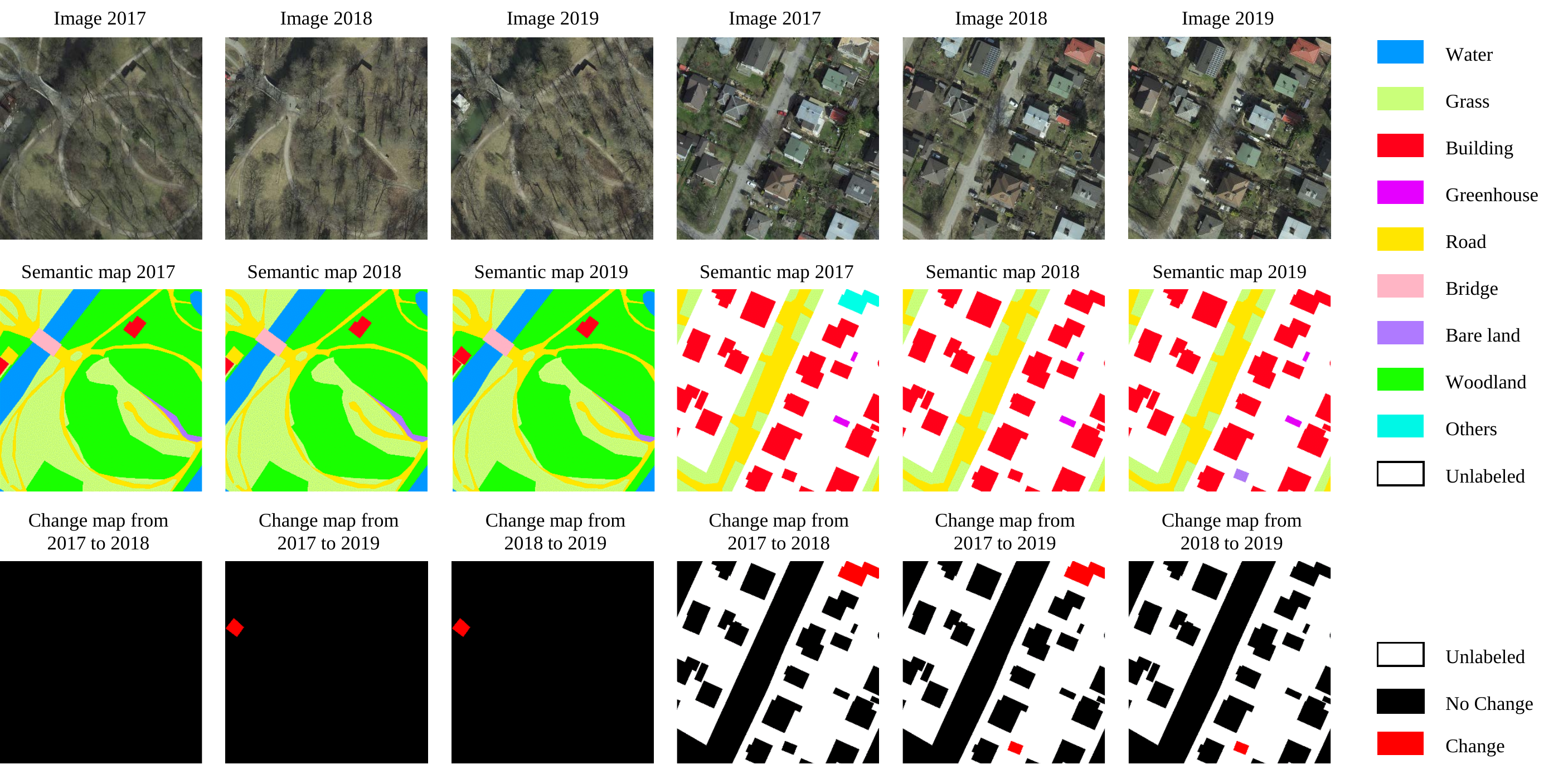}
  \caption{Examples of Hi-UCD dataset for images, semantic maps and change maps.}\label{fig:example}
\end{figure}
 In Figure \ref*{fig:example}, examples of the Hi-UCD dataset are given. Compared with other public datasets, its characteristics are as follows:

 \begin{enumerate}
   \item \textbf{Ultra-high spatial resolution.}  
   Hi-UCD uses aerial images with a spatial resolution of 0.1m, which is the highest resolution in public data. In these images, the geometric shape of the ground objects is clear, and the boundary is obvious, which provides rich spatial texture information. Therefore, it is conducive to detecting local changes of ground objects and realizing refined change detection.

  \item \textbf{Multi-temporal images.} Hi-UCD contains the images of the three years from 2017 to 2019, and gives the semantic annotation and change mask every two years (2017-2018, 2018-2019, 2017-2019). Changes are highly time-dependent, and multi-temporal data can provide temporal features, which helps researchers to conduct long-term serial studies and improve the temporal precision of UCD. In addition, the images of different years have undergone orthorectification without registration errors. At the same time, they were taken in the same season, which greatly reduced the influence of seasonal changes in vegetation.

  \item \textbf{Semantic annotation.} Considering typical urban objects and change-related objects, we developed Hi-UCD semantic annotation categories.
   There are 9 types of objects, including natural objects (water, grassland, woodland, bare land), artificial objects (Building, greenhouse, road, bridge), and others (change-related), basically include all types of urban land cover in Estonia. The above categories are mapped with the shapefile layer in the Estonian Topographic Database (ETD). Due to the inconsistency of the vector boundaries between different years in ETD, the buildings are based on the vector of each year, and the other classes are based on the vector of 2019. Through visual interpretation, we check the topographic shapefiles and modify them. Meanwhile, we compare images of different years to determine the relevant objects of the change and add the category “other”. Finally, the binary and multi-classes change masks generated through the semantic annotation results.
  \end{enumerate}
Because of these characteristics, Hi-UCD is full of challenges: (1) the increase in spatial resolution has aggravated the shadows and occlusions in the image.
(2) The changes in uninteresting ground objects such as cars will cause serious background noise during change detection. (3) High-rise buildings are tilted and geometrically mismatched due to different shooting angles at different times. 
(4) The number of category transitions caused by changes is much greater than the number of semantic categories, which increases the difficulty of multi-classes change detection task. In summary, Hi-UCD is far more diverse, comprehensive, and challenging.  

\section{Benchmark}

In order to establish a fair benchmark,  we evaluated the classic methods of binary and multi-class change detection
under a unified experimental setting and data division conditions.

\textbf{Methods} After decades of development, change detection methods have evolved from pixel-based direct comparison to data-driven deep learning methods \cite{Hussain.2013,Tewkesbury.2015,Shi.2020}. 
We chose different methods according to the different detection task. 
For binary change detection, these methods are the commonly used as comparison methods, including traditional methods (change vector analysis (CVA) \cite{Malila.1980}, multivariate alteration detection (MAD) \cite{Nielsen.1998}, the regularized iteratively reweighted MAD (IRMAD) \cite{Nielsen.2007}), and deep learning methods (FC\_EF \cite{CayeDaudt.2018}, FC\_Siam\_diff \cite{CayeDaudt.2018}, FC\_Siam\_diff \cite{CayeDaudt.2018}, FC\_Res\_EF \cite{CayeDaudt.2019}).
For multi-class change detection, the commonly used method is the post-classification comparison. After classifying images of different time phases through a classifier, like support vector machines [18], random forest [19], convolutional neural network [20] are compared to obtain change information. Considering the complexity of the objects in Hi-UCD, we only chose the classic semantic segmentation deep learning networks in computer vision ( Deeplab v3 \cite{chen2017rethinking.2017}, Deeplab v3+ \cite{Chen_2018_ECCV}, PSPNet \cite{Zhao_2017_CVPR} ) and  remote sensing ( FarSeg \cite{Zheng.2020b} ) for classification to obtain multi-class changes.

\textbf{Settings} We used 300 pairs of images in 2017 and 2018 for training, the remaining 59 pairs as the validation set, and 386 pairs of images in 2018 and 2019 for testing.
In the traditional method, the clustering method proposed in \cite{Celik.2009} was used to obtain the change mask. For all deep learning methods, the learning rate was 0.01 and use a polynomial decay with a decay factor of 0.9. The batch size was 4 and trained on a single GPU.  The stochastic gradient descent (SGD) was used for optimization with weight decay of 0.0001 and a momentum of 0.9. For data augmentation, horizontal and vertical flip, rotation of 90 degrees and random cut ( $size=(512,512)$) were adopted during training. 
In binary change detection, the number of iterations is 10k, and the loss function is the binary cross-entropy and dice loss. While in classification,  we used the cross-entropy loss function with 20k iterations. The backbone used for classification methods was ResNet-50, which was pre-trained on ImageNet \cite{Deng.2009}. 

\textbf{Metrics}
 We used overall accuracy (OA), kappa coefficient to evaluate the overall performance of the change detection results. For binary change detection methods, we used intersection over union (IoU) to only evaluate the ability to detect changes. In addition, We added mean intersection over union (mIoU) to evaluate algorithm performance in classification and multi-class change detection. The parameters and number of operations measured
 by multiply-adds (MAdd) calculated by a tensor with a size of $1\times C\times256\times256 (C=3,6)$ are given to show deep learning model complexity.
 The accuracy evaluation results of different methods are shown in Table~\ref{tab:result}.

 \textbf{Analysis} In Table~\ref{tab:firsttable},  most binary change detection methods can effectively detect unchanged ground objects, the IoU of change does not exceed 50\%.
 IRMAD performed the best with kappa 8\% higher than the other traditional methods. Deep learning methods are significantly higher than traditional methods in all metrics, which fully reflects the potential of deep learning in change detection. Traditional methods cannot distinguish false changes caused by shadows, occlusions, and uninteresting objects, while deep learning methods rely on their powerful learning capabilities to effectively remove background noise.
 Among them, the FC\_Siam\_diff method is slightly better than FC\_Siam\_conc in all metrics. FC\_EF improves IoU of change by nearly 7\%. After adding the residual module, FC\_EF\_Res increased by nearly 5\% and has the smallest parameters.  
 In Table~\ref{tab:secondtable}, the metrics of all methods for land cover classification in 2018 are higher than in 2019. Because the change of the ground objects leads to differences in the distribution of ground features at different times. 
 Through post-classification to get the results of multi-class change detection, there are many false alarms at the boundary of the ground objects in multi-class change results. In Table~\ref{tab:firsttable}, although Deeplab v3 obtains the best accuracy in multi-class change detection, it also has the highest computational complexity. 
 Besides, the accuracy of the change highly depends on the accuracy of the classification,
How to obtain reliable multi-class change detection results in urban areas is still a problem that needs research.

\begin{table} [tb]
  \caption{The quantitative evaluation of the baseline methods for Hi-UCD}
  \label{tab:result}
  \centering
  \subtable[ Change detection accuracy ]{
    \resizebox{0.48\linewidth}{!}{
    \begin{tabular}{ccccccccc}
      \toprule
      {Binary change detection} &
      \#Params (M)&
      MAdds (B)&
       OA (\%) &
       Kappa (\%) & 
       IoU (\%) &
        \\
       \midrule
      CVA \cite{Malila.1980} &-&-&40.79	&	3.98	&	11.51		\\
       MAD \cite{Nielsen.1998}&-&-	&	88.95	&	3.64	&	3.41	\\
       IRMAD \cite{Nielsen.2007}&-&-	&	88.08	&	11.78	&	9.38		\\  
       FC\_Siam\_conc \cite{CayeDaudt.2018}  &1.546&5.8& 91.25 	&	44.16	&	32.26 	 	 	\\
       FC\_Siam\_diff \cite{CayeDaudt.2018}&1.35&4.67 &91.74 	&	47.67  	&	35.19 	 	 	\\
       FC\_Siam\_EF \cite{CayeDaudt.2018}&1.35&3.54 &91.50 	&	54.92 		&	42.51 	 	 	 	\\
        Siam\_Res\_EF \cite{CayeDaudt.2019}& 1.104&1.98&93.05 	&	60.67 	&	47.62  	 	 \\
       \midrule
       { Multi-class change detection} &
       \#Params (M)&
       MAdds (B)&
         OA (\%) &
         Kappa (\%) & 
         mIoU (\%) \\
         \midrule
   
       Deeplab v3 \cite{chen2017rethinking.2017}&39.046&80.58& 76.82	&	29.48	&	17.51\\
       Deeplab v3+ \cite{Chen_2018_ECCV}&39.897&26.34&75.83	&	28.31	&	15.65\\
       PSPNet \cite{Zhao_2017_CVPR}&46.588&88.60	& 76.17	&	28.49	&	14.29\\
       FarSeg \cite{Zheng.2020b}&33.881&28.68& 73.58	&	25.68	&	13.34\\
        \bottomrule	
         \end{tabular}}
         \label{tab:firsttable}
  }
  \hfill
  \subtable[Land cover accuracy ]{  
    \resizebox{0.48\linewidth}{!}{      
    \begin{tabular}{ccccccc}
      \toprule
       Methods&
       \#Params (M)&
       MAdds (B)&
       Year&
        OA (\%) &
        Kappa (\%) & 
        mIoU (\%) 
        \\
        \midrule
        \multirow{2}{*}{Deeplab v3 \cite{chen2017rethinking.2017}}
        &\multirow{2}{*}{39.046}&\multirow{2}{*}{40.29}&2018	&	87.59	& 83.94 &	72.39 \\
        &&&2019&77.19&69.74&42.55\\
        \midrule
        \multirow{2}{*}{Deeplab v3+ \cite{Chen_2018_ECCV}} &\multirow{2}{*}{39.897 }&\multirow{2}{*}{13.17}& 2018	&	86.28	& 82.22&	67.24 \\
        &&&2019&76.23&68.45&37.98\\
        \midrule
        \multirow{2}{*}{PSPNet \cite{Zhao_2017_CVPR}} &\multirow{2}{*}{46.588 }&\multirow{2}{*}{44.30} &2018	&	86.50	&82.50&69.88 \\
        &&&2019&74.50&65.79&37.37\\
        \midrule
        \multirow{2}{*}{FarSeg \cite{Zheng.2020b}} &\multirow{2}{*}{33.881 }&\multirow{2}{*}{14.34 }& 2018	&	86.78	&82.88 &69.98 \\
        &&&2019&75.58&64.83&36.02\\
        \bottomrule
         \end{tabular}}
         \label{tab:secondtable}
  }
  \end{table}

\section{Conclusion}
In this article, we introduce a new multi-temporal ultra-high-resolution aerial image UCD dataset, which has rich semantic annotations to detect more details of urban change.
At the same time, we have established a benchmark for UCD in binary and multi-class change detection tasks.
In the next work, we will continue to expand the scale of the dataset and provide different large-area test sets to verify the generalization and migration of the algorithm better. We hope the release of Hi-UCD will promote the development of UCD.

\medskip

\ack
This work was supported by National Natural Science Foundation of China under Grant Nos.41771385, 41801267 and 42071350. The authors would like to thank the Estonian Land Board for acquiring and providing the data used in this study.
\small

\bibliography{neurips_2020}

\begin{thebibliography}{10}
\expandafter\ifx\csname url\endcsname\relax
  \def\url#1{\texttt{#1}}\fi
\expandafter\ifx\csname urlprefix\endcsname\relax\def\urlprefix{URL }\fi
\expandafter\ifx\csname href\endcsname\relax
  \def\href#1#2{#2} \def\path#1{#1}\fi

\bibitem{Hussain.2013}
M.~Hussain, D.~Chen, A.~Cheng, H.~Wei, D.~Stanley, Change detection from
  remotely sensed images: From pixel-based to object-based approaches, ISPRS
  Journal of Photogrammetry and Remote Sensing 80 (2013) 91--106.

\bibitem{Huang.2017}
X.~Huang, D.~Wen, J.~Li, R.~Qin, Multi-level monitoring of subtle urban changes
  for the megacities of china using high-resolution multi-view satellite
  imagery, Remote Sensing of Environment 196 (2017) 56--75.

\bibitem{Daudt.2018}
R.~C. Daudt, B.~{Le Saux}, A.~Boulch, Y.~Gousseau, Urban change detection for
  multispectral earth observation using convolutional neural networks, in:
  IGARSS 2018 - 2018 IEEE International Geoscience and Remote Sensing
  Symposium, IEEE, 2018, pp. 2115--2118.

\bibitem{Zhang.2020}
M.~Zhang, W.~Shi, A feature difference convolutional neural network-based
  change detection method, IEEE Transactions on Geoscience and Remote Sensing
  (2020) 1--15.

\bibitem{Benedek.2009}
C.~Benedek, T.~Sziranyi, Change detection in optical aerial images by a
  multilayer conditional mixed markov model, IEEE Transactions on Geoscience
  and Remote Sensing 47~(10) (2009) 3416--3430.

\bibitem{Fujita.2017}
A.~Fujita, K.~Sakurada, T.~Imaizumi, R.~Ito, S.~Hikosaka, R.~Nakamura, Damage
  detection from aerial images via convolutional neural networks, in: IAPR
  MVA2017, {MVA Organization}, 2017, pp. 5--8.

\bibitem{Chen.2020}
H.~Chen, Z.~Shi, A spatial-temporal attention-based method and a new dataset
  for remote sensing image change detection, Remote Sensing 12~(10) (2020)
  1662.

\bibitem{Ji.2019b}
S.~Ji, S.~Wei, M.~Lu, Fully convolutional networks for multisource building
  extraction from an open aerial and satellite imagery data set, IEEE
  Transactions on Geoscience and Remote Sensing 57~(1) (2019) 574--586.

\bibitem{Lebedev.2018}
M.~A. Lebedev, Y.~V. Vizilter, O.~V. Vygolov, V.~A. Knyaz, A.~Y. Rubis, Change
  detection in remote sensing images using conditional adversarial networks,
  ISPRS - International Archives of the Photogrammetry, Remote Sensing and
  Spatial Information Sciences XLII-2 (2018) 565--571.

\bibitem{CayeDaudt.2019}
R.~C. Daudt, B.~{Le Saux}, A.~Boulch, Y.~Gousseau, Multitask learning for
  large-scale semantic change detection, Computer Vision and Image
  Understanding 187 (2019) 102783.

\bibitem{Bourdis.2011}
N.~Bourdis, D.~Marraud, H.~Sahbi, Constrained optical flow for aerial image
  change detection, in: I.~Staff (Ed.), 2011 IEEE International Geoscience and
  Remote Sensing Symposium, IEEE, 2011, pp. 4176--4179.

\bibitem{Tewkesbury.2015}
A.~P. Tewkesbury, A.~J. Comber, N.~J. Tate, A.~Lamb, P.~F. Fisher, A critical
  synthesis of remotely sensed optical image change detection techniques,
  Remote Sensing of Environment 160 (2015) 1--14.

\bibitem{Shi.2020}
W.~Shi, M.~Zhang, R.~Zhang, S.~Chen, Z.~Zhan, Change detection based on
  artificial intelligence: State-of-the-art and challenges, Remote Sensing
  12~(10) (2020) 1688.

\bibitem{Malila.1980}
W.~Malila, Change vector analysis: An approach for detecting forest changes
  with landsat, LARS Symposia (1980).

\bibitem{Nielsen.1998}
A.~A. Nielsen, K.~Conradsen, J.~J. Simpson, Multivariate alteration detection
  (mad) and maf postprocessing in multispectral, bitemporal image data: New
  approaches to change detection studies, Remote Sensing of Environment 64~(1)
  (1998) 1--19.

\bibitem{Nielsen.2007}
A.~A. Nielsen, The regularized iteratively reweighted mad method for change
  detection in multi- and hyperspectral data, IEEE transactions on image
  processing : a publication of the IEEE Signal Processing Society 16~(2)
  (2007) 463--478.

\bibitem{CayeDaudt.2018}
R.~C. Daudt, B.~{Le Saux}, A.~Boulch, Fully convolutional siamese networks for
  change detection, in: I.~I. C. o.~I. Processing (Ed.), 2018 IEEE
  International Conference on Image Processing, IEEE, 2018, pp. 4063--4067.

\bibitem{chen2017rethinking.2017}
L.-C. Chen, G.~Papandreou, F.~Schroff, H.~Adam, Rethinking atrous convolution
  for semantic image segmentation, arXiv preprint arXiv:1706.05587 (2017).

\bibitem{Chen_2018_ECCV}
L.-C. Chen, Y.~Zhu, G.~Papandreou, F.~Schroff, H.~Adam, Encoder-decoder with
  atrous separable convolution for semantic image segmentation, in: The
  European Conference on Computer Vision (ECCV), 2018.

\bibitem{Zhao_2017_CVPR}
H.~Zhao, J.~Shi, X.~Qi, X.~Wang, J.~Jia, Pyramid scene parsing network, in: The
  IEEE Conference on Computer Vision and Pattern Recognition (CVPR), 2017.

\bibitem{Zheng.2020b}
Z.~Zheng, Y.~Zhong, J.~Wang, A.~Ma, Foreground-aware relation network for
  geospatial object segmentation in high spatial resolution remote sensing
  imagery, in: The IEEE/CVF Conference on Computer Vision and Pattern
  Recognition (CVPR), 2020, pp. 4096--4105.

\bibitem{Celik.2009}
T.~Celik, Unsupervised change detection in satellite images using principal
  component analysis and $k$-means clustering, IEEE Geoscience and Remote
  Sensing Letters 6~(4) (2009) 772--776.

\bibitem{Deng.2009}
J.~Deng, W.~Dong, R.~Socher, L.-J. Li, {Kai Li}, {Li Fei-Fei}, Imagenet: A
  large-scale hierarchical image database (2009).

\end{thebibliography}
\bibliographystyle{elsarticle-num}

\end{document}